\pdfoutput=1

\documentclass[11pt]{article}

\usepackage{acl}

\usepackage{times}
\usepackage{latexsym}
\usepackage{graphicx}
\usepackage[T1]{fontenc}

\usepackage[utf8]{inputenc}

\usepackage{microtype}

\title{A Multilingual Evaluation of NER Robustness to Adversarial Inputs} 
\author{Akshay Srinivasan \\
  University of Ottawa, Canada \thanks{Work done during an internship at National Research Council, Canada}\\
  \texttt{asrin033@uottawa.ca} \\\And
  Sowmya Vajjala \\
  National Research Council, Canada\\
  \texttt{sowmya.vajjala@nrc-cnrc.gc.ca} \\}

\begin{document}
\maketitle
\begin{abstract} 
Adversarial evaluations of language models typically focus on English alone. In this paper, we performed a multilingual evaluation of Named Entity Recognition (NER) in terms of its robustness to small perturbations in the input. Our results showed the NER models we explored across three languages (English, German and Hindi) are not very robust to such changes, as indicated by the fluctuations in the overall F1 score as well as in a more fine-grained evaluation. With that knowledge, we further explored whether it is possible to improve the existing NER models using a part of the generated adversarial data sets as augmented training data to train a new NER model or as fine-tuning data to adapt an existing NER model. Our results showed that both these approaches improve performance on the original as well as adversarial test sets. While there is no significant difference between the two approaches for English, re-training is significantly better than fine-tuning for German and Hindi.  
\end{abstract}

\section{Introduction}
\label{sec:intro}
NLP systems are traditionally evaluated and compared against a gold standard, which is generally immutable. Recent research has shown that even the NLP systems that perform well on the standard test set show a significant drop in performance even for small perturbations in the input test data, across a range of NLP tasks \cite{gardner-etal-2020-evaluating}. Although this strand of research covered many tasks, it has been exclusively focused on English, with a few exceptions (e.g., \newcite{Shmidman.Guedalia.ea-20}, for Hebrew). Further, to our knowledge, the primary usage of such adversarial test sets have been in either evaluating NLP models or in usage as additional, augmented data to improve model robustness, without much focus on using the new data for fine-tuning, instead of re-training. A better understanding of fine-tuning with adversarial test sets is important, and useful in most real-world scenarios, where we may not have access to the original training data while having access to the trained model itself.  

Named Entity Recognition (NER) is among the most common NLP tasks both in research and in industry applications \cite{Lorica.Nathan-21}. Although much progress has been made on NER over the past decades, existing NER systems were also shown to be sensitive to small changes in input data in the past \cite{lin-etal-2021-rockner,vajjala-balasubramaniam:2022:LREC}. Table~\ref{tab:tabexample1} shows an example of how predictions can change with minor changes in input, for one of the state of the art NER models\footnote{\url{https://huggingface.co/flair/ner-english}}. Going by the original sentence, all three sentences should carry the \verb|LOC| tag for the entity in the sentence. However, that is not the case, as the outputs shows. Clearly, small, and seemingly harmless changes are changing model predictions. 

\begin{table}[htb!]
\centering
\begin{tabular}{|p{6.5cm}|}
\hline Original: It was the second costly blunder by \textbf{Syria\_LOC} in four minutes . \\
\hline Altered: It was the second costly blunder by \textbf{Hyderabad\_ORG} in four minutes . \\
\hline Altered: It was the second costly blunder by \textbf{Hyderabad\_LOC} in four hours . \\
\hline
\end{tabular}
\caption{Illustration of an NER model's predictions with minor changes to an original test set sentence}
\label{tab:tabexample1}
\end{table} 

Even recent large language models such as ChatGPT struggle with sequence tagging tasks such as NER, across multiple languages\cite{qin2023chatgpt,lai2023chatgpt,wu2023bloomberggpt}, which clearly illustrates that NER is far from being considered solved. In this backdrop, considering the significance of NER in research and practical scenarios, a better understanding of how a model's predictions change with slight changes in input becomes an important issue to address. Hence, we explore the following questions in this paper:
\begin{enumerate}
    \item How does the performance of NER models across three languages change with small changes to the original input?
    \item How does retraining an NER model with adversarial data augmentation compare with adversarial fine-tuning of NER across languages?
\end{enumerate} 

Our contributions are summarized as follows:
\begin{itemize}
    \item We conducted the first comparative study of the robustness of NER models beyond English, covering three languages, in a space where all previous work focused on English alone. 
    \item We report first results on the comparison between data augmentation and adversarial fine-tuning for NER, for all the three languages. 
    \item We show how existing methods for data augmentation can be repurposed to develop language-agnostic methods to generate adversarial test sets for NER. 
\end{itemize}

Starting with a conceptual background (Section~\ref{sec:relw}), we describe our methods for adversarial dataset creation (Section~\ref{sec:advgen}) and the general experimental setup (Section~\ref{sec:setup}) followed by a detailed discussion of our results (Section~\ref{sec:results}) and a summary (Section~\ref{sec:concl}), focusing on the ethical impacts, limitations and broader impact towards the end. 

\section{Related Work}
\label{sec:relw}
Evaluating using multiple datasets is one of the ways to assess the robustness and generalization capabilities of NLP models. Developing challenge sets, and generating adversarial datasets that can potentially cause a model to fail, are some possibilities in this direction \cite{isabelle-etal-2017-challenge,ettinger-etal-2017-towards,glockner-etal-2018-breaking,gardner-etal-2020-evaluating}.
Adversarial data generation in NLP focuses on surface-level perturbations to the input text, proposing various means of insertion/deletion/swapping of words/characters/sentences \cite{jia-liang-2017-adversarial,gao2018black,ribeiro-etal-2018-semantically}. Other approaches such as paraphrasing \cite{iyyer-etal-2018-adversarial}, generating semantically similar text using other deep learning models \cite{zhao2018generating,michel-etal-2019-evaluation}, using a human-in-the-loop \cite{wallace-etal-2019-trick} were also explored in the past. While many of the proposed methods are black-box approaches, assuming no knowledge about the NLP models themselves, some of the approaches are white box, with more access to the inner workings of a model \cite{liang2018deep,blohm-etal-2018-comparing,wallace-etal-2019-universal}, and some models implement both \cite{li2019textbugger}. We focus on one specific NLP task - NER, and only work on black-box methods in this paper.  

In terms of the strategies to protect models against adversarial attacks, the most common approach followed by past NLP research has been to incorporate adversarial data into the training process, through data augmentation, or using adversarial training as a regularization method (See \newcite{goyal2022survey} and \newcite{zhang2020adversarial} for a detailed overview). Using adversarial data for full re-training of a model (as is the case with data augmentation) assumes access to the original data and the model, which is not practical in many real-world scenarios. One possibility to explore in such cases is to test whether using adversarial data to fine-tune a trained NER model improves its robustness. We compare using the adversarial data (through data augmentation) for re-training an NER model versus using it only for fine-tuning a previously trained NER model in this paper.

\paragraph{Adversarial Testing and Data Augmentation in NER:} Adversarial testing approaches for NER in the previous work was entirely done for English datasets, and primarily focused on methods replace entities in the original test set with new ones using gazetteers or other means \cite{Agarwal.Yang.ea-20,vajjala-balasubramaniam:2022:LREC}. \newcite{lin-etal-2021-rockner} used entity linking and masked language models coupled with an existing NER model to generate adversarial test sets for NER. More recently, \newcite{das-paik-2022-resilience} used grammatical case information to generate adversarial test sets for NER. \newcite{simoncini-spanakis-2021-seqattack} proposed other ways to make small changes to the context around entities in a sentence, to generate adversarial test sets. Other related work \cite{mathew2019biomedical,ding-etal-2020-daga, zhu2021improving} focused on data augmentation for NER, that require training of new, additional models. While we use our adversarial datasets for data augmentation too later in the paper, the novelty of the current research, compared to this existing body of work focusing on NER, comes in two forms:
\begin{enumerate}
    \item While all previous work exclusively focused on English NER so far, we perform experiments with three languages - English, German and Hindi.
    \item The approaches we used are lightweight, language agnostic means to generated adversarial test sets for NER, which do not rely on the availability of additional tools like entity linkers, and do not also need any additional training to generate the datasets. 
\end{enumerate} 



\section{Adversarial Test Set Creation}
\label{sec:advgen}
Our adversarial dataset creation methods can be broadly classified into two approaches - replacing entities and changing contexts. All except one method work for all the three languages we tested, and can be easily expanded to add other languages. Relevant code and generated datasets are provided as supplementary material\footnote{\url{https://dx.doi.org/10.6084/m9.figshare.22674079}}. 
\subsection{Replacing Entities}
We implemented two methods that replace the entity occurrences in the test set with another entity of the same category, keeping the rest of the sentence unchanged. Thus, they don't change the grammatical structure of the sentence and tell us  how much the NER systems learn beyond memorizing the entities. 

\paragraph{Random Sampling (\textit{RS})}
All the entity occurrences of the same type are shuffled throughout the test set in this approach. This is a simple and easily portable method across languages, which could serve as a strong baseline. 

\paragraph{Gazetteers (\textit{Faker})}
This approach replaces existing entities with new entities of the same type using an existing gazetteer. Faker\footnote{\url{https://faker.readthedocs.io/}} is a python library that generates fake data for various application purposes, which supports multiple languages, and regions. We used it to replace the \textit{Person} and \textit{Location} entities in all the datasets, in all three languages we experimented with. \newcite{vajjala-balasubramaniam:2022:LREC} used Faker for adversarial NER test sets, but they used only English (on OntoNotes dataset). We randomly choose among three locale settings for English (USA, Canada, India) and German (Germany, Austria, and Switzerland) respectively for replacement. For Hindi, there was only one locale setting provided (HI-IN). 

\subsection{Changing the Context}
These approaches deal with changing the context in which the entities occur in a sentence by making small changes to the tokens around it. 

\paragraph{Masking (\textit{Mask})}
We leveraged transformer based pre-trained language models trained with a masked language modeling objective to change the context in the original test datasets.  We masked up to three randomly chosen non-entity tokens per sentence, and used the language model to generate those tokens, thereby creating new sentences with the same entities, but slightly altered contexts. Since there are multilingual pre-trained language models available, this approach is applicable to multiple languages. 

\begin{table*}[htb!]
\centering
\begin{tabular}{|l|p{14cm}|}
\hline
\textbf{Test set} & \textbf{Sentence} \\
\hline
\verb|Orig| & "We suspect that these killings are linked to politics,” spokesman Bala Naidoo told Reuters. \\
\verb|RS| & “We suspect that these killings are linked to politics,” spokesman Deborah Compagnoni told Watford. \\
\verb|Faker| & “We suspect that these killings are linked to politics,” spokesman Jeremy Shukla told Reuters. \\
\verb|Mask| & “We suspect the these killings are linked to politics,” spokesman Bala Naidoo tells Reuters,\\
\verb|Para| & “We assume that these killings are political in nature”, spokesman Bala Naidoo told Reuters.\\
\verb|M+R| & now we suspect that these killings are connected in politics, now spokesman Deborah Compagnoni told Watford.\\
\hline
\end{tabular}
\caption{Adversarial variations generated for a test sentence from conll03-En}
\label{tab:tabexample}
\end{table*} 

\paragraph{Paraphrasing (\textit{Para})}
The objective behind this approach is to alter the structure of the input sentences, while keeping the named entities intact. About 500 sentences were randomly chosen from the test set and fed to an online, subscription based english paraphraser, Quillbot\footnote{\url{https://quillbot.com/}}, which was shown to generate better paraphrases than other approaches such as back-translation or using GPT-3 in recent research \cite{shiri.zhuo.ea-22} and in our initial evaluations. The paraphrased output obtained for each sentence is then taken and the entity tokens from the original test set are mapped to the entity tokens of the paraphrased sentences with the respective entity tags, leaving the rest of the tokens as \verb|O|. Limiting to 500 sentences is primarily due to the fact that Quillbot did not provide an API, and there is a limit to the amount of text one can paraphrase per request, even with subscription.

There are some challenges with this approach, though. While Quillbot's paraphrases when we choose the "Fluency" setting are of good quality, and are always grammatically correct, they sometimes alter the entities themselves (e.g., United States can become U.S. in the paraphrased version) or change the original tokenization of the dataset. In such cases where an automatic mapping between the entity tags from the original sentence and tokens of the paraphrased sentence failed, we discarded the sentence from our test set. Note that this approach is compatible only with English and we aren't aware of any reliable paraphrasers for other languages. To our knowledge, full paraphrasing wasn't used for adversarial test sets in NER before. A total of 399 sentences from conll03-en, 373 sentences from mconer21-en, and 378 sentences from wnut17 were finally used as test sets. 

\subsection{Changing Entity + Context}
\paragraph{Masking + Random Sampling (\textit{M+R})}
All the previous approaches focused on either trying to alter the context alone or replace the entities alone. This approach, does both by combining masking with random sampling. Since both these approaches are straightforward and can work across languages, they can be combined to create a new adversarial test set in all three languages. Table~\ref{tab:tabexample} shows one example of how the various approaches alter a single sentence from one of the English datasets. As with most data augmentation or adversarial generation approaches in NLP, some of the generated text may contain minor grammatical errors, as seen in in \verb|Mask| and \verb|M+R| settings. However, Compared to other common approaches such as those that involve insertion/deletion/swapping of words/characters or back-translation, the potential errors introduced by masking alone are minor. Further even the original datasets themselves contain sentences with such minor errors. So, we don't foresee this affecting the main findings of the paper. A publicly available NER model \footnote{\url{https://huggingface.co/flair/ner-english}} predicts correctly for all these sentences. 

\section{Experimental Setup}
\label{sec:setup}
We experimented with three English, two German, and one Hindi NER datasets \footnote{While it is theoretically possible to add more languages to this paper, the choice of these languages is motivated by the authors' familiarity with the languages, which is essential for qualitative analysis}. The first half of our experiments focused on testing the NER models on adversarial test sets, following which we compared the effects of adversarial fine-tuning and data augmentation on the performance of the NER models. All the experiments were carried out on a high performing computing resource that ran an Nvidia A100 GPU with 32 GB RAM. Further details on the experimental setup are as follows.

\subsection{Datasets}
NER datasets from three popular shared tasks -  conll03 \cite{tjong-kim-sang-de-meulder-2003-introduction}, multiconer21 \cite{malmasi-etal-2022-semeval} and wnut17 \cite{derczynski-etal-2017-results} were considered and their corresponding language subsets for English (conll03-en, mconer21-en, wnut17), German (conll03-de, mconer21-de) and Hindi (mconer21-hi) were used. Conll03 datasets have a tag set with four entity types (PER, LOC, ORG and MISC), and multiconer21 and wnut17 share the tag set consisting of six entity types (PER, LOC, CORP, GRP, CREATIVE-WORK and PROD). While the sentences in conll03 came from news articles, multiconer21 was collected from three domains (wikipedia sentences, questions, search queries), and wnut17 consisted of sentences from social media sources such as twitter, youtube and reddit.\footnote{Some statistics about the datasets are shown in Appendix~\ref{sec:app-ds}.} We created five adversarial test sets for each of the three English datasets, four adversarial sets for each of the two German datasets and the Hindi dataset respectively, resulting in a total of 27 adversarial test sets covering three languages and six datasets.

\subsection{NER Models}
We used a combination of existing state of the art NER models (if available) and fine-tuning a pre-trained language model for all the languages/datasets. They are explained below.

\paragraph{TNER and Fine-tuned BERT: } TNER\footnote{\url{https://github.com/asahi417/tner}} is a Python library to train transformer based language models for NER tasks \cite{Ushio.Camacho-Collados-21}, which has pre-trained for mconer21-en, wnut17-en and conll03-en datasets. We use these models to perform our experiments for English NER. We will refer to this approach as \textit{tner}. While TNER's model hub had publicly available models for other languages as well, the performance of the models was lower compared to the state of the art, and hence, we fine-tuned the multilingual BERT \cite{devlin2019bert} model hosted on Huggingface\footnote{\url{https://huggingface.co/bert-base-multilingual-uncased}} for NER on German (conll03-de, mconer21-de) and Hindi (mconer21-hi) datasets. We will refer to this approach as \textit{mbertft}. 

We followed the same approach for the later re-training (i.e., training the NER model again using the original training data augmented with adversarial data) and fine-tuning (using adversarial data is used only to fine-tune the existing NER model) experiments in Section~\ref{subsec:advtuning}, and report the results with \textit{tner} for English and \textit{mbertft} for German and Hindi, as this setup gave the best results even in those experiments. For adversarial fine-tuning, \textit{tner} and \textit{mbertft} were fine-tuned for 4 epochs, learning rate was set to 0.0001 and the batch size was set to 16. For the Adversarial re-training however, the models were trained for 6 epochs since it is being trained from scratch, while the other hyper parameters were kept the same. The hyperparameter settings were from the original BERT paper \cite{devlin2019bert}. We used 60\% of the adversarial test set for data augmentation+training/fine-tuning and used the remaining 40\% to test the approaches. For both adversarial fine-tuning and re-training, we report the average F1 score over 10 runs, with different random seeds, and compare them in terms of statistical significance using a paired t-test. 

\paragraph{Stanza: } Stanza \cite{qi2020stanza} is a Python based NLP toolkit that hosts a few pre-trained NER models trained with a BiLSTM+CRF architecture. We evaluated Stanza's pre-trained conll-en and conll-de models using our generated adversarial test sets. 

\paragraph{Flair: } Flair is a popular NLP library that is widely used for performing NLP tasks \cite{akbik2019flair}. We evaluated the pre-trained NER models provided by Flair for conll-en and conll-de.

There are other NER models that offer slightly better performance than Flair/Stanza/BERT-fine tuning, and there are other large language models to explore, but we focused on publicly available/downloadable models for NER and and easily re-implementatble benchmarks (e.g., BERT finetuning) in this paper. It would be interesting to extend this to cover additional methods in future, but we limit to a smaller set of models to maintain a manageable number of experiments and do a meaningful analysis later. 

\subsection{Evaluation}
Micro-F1 score from seqeval \cite{seqeval} was used as the evaluation metric to test the robustness of the NER models, as it is the most commonly reported measure for this task.

\paragraph{Nervaluate:} Nervaluate\footnote{\url{https://github.com/MantisAI/nervaluate}} is a Python library for performing a more fine-grained evaluation of NER, and is based on the metrics from a SemEval 2013 task \cite{segura-bedmar-etal-2013-semeval}. Apart from giving a single F1 score, it calculates the efficiency of the model using five error categories: correct, incorrect, partially correct, missing labels (an entity tagged as non-entity) and spurious labels(a non-entity tagged as entity). The error metrics are reported in four formats: strict (both entity span and entity type match), exact (entity span matches, irrespective of the type), partial (partial span match, irrespective of the type), and type (some overlap between gold annotation and system prediction). We used nervaluate to compare the performance of NER on the original and adversarial test sets, to understand what kind of errors affect their performance. 
 
\section{Results}
\label{sec:results}
Our experiments aimed at understanding the robustness of NER models to adversarial test sets and exploring whether adversarial data augmentation and fine tuning will help boost the performance on adversarial test sets. The results of these experiments are discussed below.

\subsection{Adversarial Testing} Adversarial test sets were created by implementing the approaches mentioned in Section~\ref{sec:advgen} and tested on pre-trained NER models for all the three English datasets, and for conll03-de. As described earlier, we also fine-tuned a multilingual bert model for German and Hindi datasets. Tables \ref{tab:tab1}, \ref{tab:tab2} and \ref{tab:tab3} summarize the performance of all the NER models we tested on, for English, German and Hindi respectively, in terms of the micro-F1 score.

\begin{table}[htb!]
\centering
\begin{tabular}{|l|p{0.5cm}p{0.6cm}p{0.5cm}|l|l|}
\hline
& \multicolumn{3}{c|}{\textbf{conll03-en}} & \textbf{wnut17} & \textbf{mconer21}\\
\hline
test set& tner & stanza & flair & tner & tner \\ \hline
\verb|Orig| & 0.91 &0.92& 0.92& 0.60 & 0.81 \\
\verb|RS| & 0.87 &0.89&0.89& 0.63 & 0.76\\
\verb|Faker| & 0.84 &0.85&0.86& 0.64 & 0.79\\
\verb|Mask| & 0.84 &0.85&0.85& 0.58 & 0.75\\
\verb|Para| & 0.80 &0.72&0.78& 0.65 & 0.64\\
\verb|M+R| & 0.82 &0.85&0.83& 0.53 & 0.77\\
\hline
\end{tabular}
\caption{English NER performance (Micro-F1)}
\label{tab:tab1}
\end{table}

\begin{table}[htb!]
\centering
\begin{tabular}{|l|ccc|c|}
\hline
\textbf{} & \multicolumn{3}{c|}{\textbf{conll03-de}} & \textbf{mconer21}\\
\hline
test set & mbertft & stanza & flair & mbertft \\ \hline
\verb|Orig| & 0.83 &0.85&0.83 & 0.70\\
\verb|RS| & 0.80 &0.82&0.78& 0.58\\
\verb|Faker| & 0.81 &0.85&0.81& 0.55\\
\verb|Mask| & 0.78 & 0.81&0.81& 0.53\\
\verb|M+R| & 0.75 &0.80&0.76& 0.50\\
\hline
\end{tabular}
\caption{German NER performance (Micro-F1)}
\label{tab:tab2}
\end{table}

\begin{table}
\centering
\begin{tabular}{|lc|}
\hline
\textbf{Test set} & \textbf{mconer21-hi}\\
\hline
\verb|Orig| & 0.62 \\
\verb|RS| & 0.55 \\
\verb|Faker| & 0.61 \\
\verb|Mask| & 0.52 \\
\verb|M+R| & 0.48 \\
\hline
\end{tabular}
\caption{Hindi NER performance (Micro-F1)}
\label{tab:tab3}
\end{table}

For English NER, results from Table \ref{tab:tab1} show a clear drop in NER model performance for two out of the three datasets (conll03 and mconer21), with the largest drop seen in the test set obtained by paraphrasing using Quillbot. However, it is important to note that the size of the test set for paraphrasing is far smaller (399, 373 and 378 sentences respectively for conll-03, mconer21 and wnut17) than the original test set, as explained earlier in Section~\ref{sec:setup}. So, the drop is not directly comparable with other test sets which are larger in size.  

Apart from this, there are also differences among  individual methods for all the datasets. For example, \verb|Faker| dataset appears to have had a stronger effect on all the three models trained on conll03-en dataset compared to multiconer-en dataset. Considering that there is generally similar drop across all three models of conll03-en, we would speculate that the difference in performance is due to the differences in the dataset composition and entity categories. 

For wnut17, the drop is largest for \verb|M+R|, followed by \verb|Mask|. Considering that only those test sets involving masking resulted in a drop for this dataset, we could safely attribute this drop to the difference in the nature of the data that the masked language model was exposed to, compared to the very noisy social media data in wnut17. An interesting aspect of testing with wnut17 model is that the model's performance was better on 3 out of 5 adversarial test sets, compared to the original test set. We believe it is important to note in this context that the NER model for wnut17 also has the lowest performance on the original test set among the three datasets and wnut17 is the noisiest data of them all (social media content). Considering that a model with a much lower overall F1 score, and trained on the most noisy dataset among the three, was still relatively more robust to three of the adversarial test sets, future research should perhaps also take a closer look at other, additional means of evaluating NER models instead of using the standard F1 score as the sole criterion to choose the best model, as was also suggested by other recent research on the topic. \cite{vajjala-balasubramaniam:2022:LREC}.  

For both German and Hindi NER (Table \ref{tab:tab2} and Table  \ref{tab:tab3}), we observe that the drop is the highest for masking+random sampling datasets in both the languages. Between the two German datasets, the drop in f1 scores for adversarial datasets appear to be much larger for mconer21 than conll03. A possible reason could be the poorer performance of the original mconer21 model itself. 

While there are several other interesting results to compare and discuss across languages and datasets, overall, these results indicate that the NER models are not fully robust when tested against new test sets created with easily replicable, generally language-agnostic approaches. Their performance drop is larger when context altering approaches are employed. One question we asked ourselves at this point is: what exactly are the adversarial test sets changing in the model performance?

\subsection{Fine-Grained Evaluation}
\begin{figure*}[htb!]
\centering
  \includegraphics[width=\linewidth]{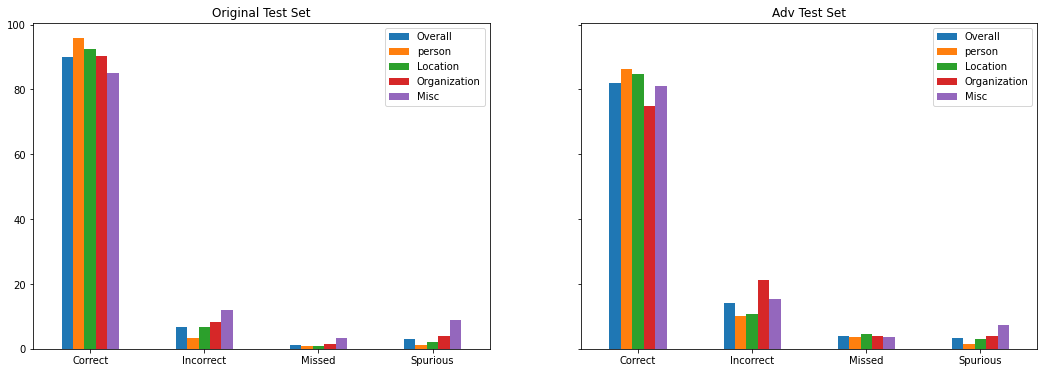}
  \caption{The performance (micro-F1) of an English NER model on original and M+R adversarial test sets}
  \label{fig:plot}
\end{figure*}

We used nervaluate to understand what aspect of NER performance is mainly affected by the adversarial data. Since there are many models, train and test sets, we choose one train/test set and model combination for this analysis. Figure~\ref{fig:plot} shows the analysis for TNER's pre-trained NER model trained on conll03-en dataset, as a comparison between the original test set and the \verb|M+R| adversarial test set (Figures~\ref{fig:plotde} and ~\ref{fig:plothi} in the appendix show the same analysis for German and Hindi respectively). 

Apart from the overall decline in performance, a closer look at the `correct' and `incorrect' categories indicate that this NER model's performance resulted in a larger drop for `organization' entity type in English. This could be because of the ambiguity involved in the entity type itself, as not every `organization' entity suits every context in which that entity type appears. Surprisingly, although the `misc' category suffers from the same problem, we don't see a large dip for that category. 

Figure~\ref{fig:plot} also indicates that there are more missed entities in the adversarial test set compared to spurious labels (non-entities tagged as one of the entity types). We compared the "strict" versus "exact" evaluation schema in nervaluate, to understand whether this increase in missed entities is a result of getting the span right, but identifying the entity type wrong. This comparison showed that while there is a 9\% drop in the overall F1 score between the original and the adversarial test sets with `strict' evaluation, there is only a 3\% drop in terms of identifying the entity spans correctly (More details in Table~\ref{tab:se-eng} in the Appendix. Tables~\ref{tab:se-deu} and ~\ref{tab:se-hi} in the appendix show this comparison for German and Hindi datasets). 

\begin{table*}[htb!]

\centering
\begin{tabular}{|p{15cm}|}
\hline \textbf{(a)} \verb|Orig|: Nicol$_{PER}$ was full of praise for his opponent who has battled testicular cancer to return to the circuit . \\
\hline \verb|RS|: Major$_{not identified}$ was full of praise for his opponent who has battled testicular cancer to return to the circuit . \\ \hline
\hline \textbf{(b)} \verb|Orig|: [Nader Jokhadar]$_{PER}$ had given Syria$_{LOC}$ the lead with a well-struck header in the seventh minute . \\
\verb|Faker|: [Roger Turner]$_{PER}$ had given [Timor-Leste]$_{ORG}$ the lead with a well-struck header in the seventh minute . \\ \hline \hline

\hline \textbf{(c)} \verb|Orig|: The richest parts of the property to the north and south of the central region have been estimated by \textbf{Bre-X$_{ORG}$} to contain 57 million ounces of gold . \\
 \verb|Para|: \textbf{Bre-X$_{not identified}$} estimates that the richest areas of the property to the north and south of the centre region contain 57 million ounces of gold .\\ \hline
 \hline  \textbf{(d)} \verb|Orig|: Tasmania$_{LOC}$ 352 by three ( [David Boon]$_{PER}$ 106 not out , [Shaun Young]$_{PER}$ 86 not out , [Michael DiVenuto]$_{PER}$ 119 ) v \textbf{Victoria$_{ORG}$} . \\
 \verb|Mask|: Tasmania$_{LOC}$ 352 by three ( [David Boon]$_{PER}$ 106 not out , [Shaun Young]$_{PER}$ 86 not out , [Michael DiVenuto]$_{PER}$ 119 ) v \textbf{Victoria$_{LOC}$} : \\ \hline \hline

 \hline  \textbf{(e)} \verb|Orig|:\textbf{Indianapolis$_{LOC}$} closes with games at [Kansas City]$_{LOC}$ and Cincinnati$_{LOC}$ . \\
 \verb|Mask|: \textbf{Indianapolis$_{ORG}$} begins with games at [Kansas City]$_{LOC}$ and Cincinnati$_{LOC}$ . \\ \hline 
\end{tabular}
\caption{Examples of cases where a NER model fails on adversarial instances}
\label{tab:erroranal}
\end{table*} 

\subsection{Qualitative Error Analysis} 
Apart from quantitative analysis, we also did some manual analysis to understand what kind of transformations led NER models to predict erroneous tags. Table~\ref{tab:erroranal} shows some of the correctly tagged examples taken from conll03-en test set and their adversarial counterparts with some tagging errors, using Stanza's NER model. 

When the entity `Nicole' was replaced with the entity 'Major' (Examples \textbf{a} and \textbf{b}), the first one resulted in the model missing the entity (Major not recognised as person), and the second example saw the replaced entity `Timor-Leste' (another name for the country East Timor) being mis-identified as an ORG instead of a LOC. While the transformed sentence in \textbf{c} appears very different from its source, there is only one entity, and a human reader would not find it difficult to identify the entity. However, the model missed identifying it altogether. Finally, both the masked examples (\textbf{d} and \textbf{e}) made very minor changes to the original sentence. But the model predictions changed from ORG $->$ LOC for the entity `Victoria' in a sentence, and LOC $->$ ORG for the entity `Indianapolis' in the other. In the last two cases, it can be argued that the original labels are ambiguous themselves. While that is, indeed, the case, the issue we particularly highlight is the way model predictions changed because of textual changes that should not really cause label changes. Similar trends can be observed in German and Hindi as well (Examples for German and Hindi are in the appendix in Table~\ref{tab:erroranalde} and Figure~\ref{fig:erroranalhi}). While it is definitely possible to do further qualitative analysis, we would speculate that combining this kind of analysis with explainable NLP approaches may give a more complete picture in the future on why NER model predictions fluctuate even for minimal perturbations in input text. 

\subsection{Adversarial Fine-Tuning}
\label{subsec:advtuning}
One approach to make models more robust to adversarial inputs is to include such data in building the model itself. We explored two methods in that direction: a) augmenting the training data with part of the adversarial dataset and re-training the NER model from scratch b) using adversarial data to fine-tune an existing NER model. The second method is especially useful in real-world scenarios where we have access to a trained model, but not to the original training data itself. We compared both the methods for all three languages, training one NER model per language (with conll-en, conll-de, and mconer-hi datasets respectively), and compared the performance with the \verb|M+R| test set and the original test set in each case. As mentioned in Section~\ref{sec:setup}, 60\% of the adversarial test set was used for augmented re-training/fine-tuning and the remaining 40\% was used as test data. Table \ref{tab:tab4} summarizes the results of this experiment.

\begin{table}[htb!]
\centering
\begin{tabular}{|p{2cm}p{1cm}p{1.5cm}p{1.5cm}|}
\hline
\textbf{} & \textbf{orig.} & \textbf{adv. fine-tuning}& \textbf{aug. re-training} \\

\hline
\multicolumn{4}{c}{conll-en} \\
\hline
\verb|Original| & 0.91 & 0.90 & 0.90\\
\verb|Adv Test| & 0.82 & 0.87 & 0.87\\
\hline
\multicolumn{4}{c}{conll-de} \\
\hline
\verb|Original| & 0.83 & 0.84 & 0.89$^*$\\
\verb|Adv Test| & 0.75 & 0.81 & 0.85$^*$\\
\hline
\multicolumn{4}{c}{mconer-hi} \\
\hline
\verb|Original| & 0.62 & 0.64 & 0.70$^*$\\
\verb|Adv Test| & 0.48 & 0.55 & 0.58$^*$\\
\hline
\end{tabular}
\caption{Micro-F1 score for Adversarial fine-tuning versus re-training \\ (* indicates a statistically significant difference)}
\label{tab:tab4}
\end{table}

While there are no significant differences between adversarial fine-tuning and re-training for English, and both give a 5\% performance boost on adversarial test set without compromising on the original test set performance, for both German and Hindi, re-training was significantly better than fine-tuning with both test sets (p$<$0.001). A possible reason for this difference could lie in the relatively superior performance of the original model itself. Re-training could be more useful when the performance of the original model is poor, as was the case for German and Hindi. Interestingly, just fine-tuning still improved performance on adversarial test sets by over 5\% for all languages. 

To connect these results back to our second research question (Section ~\ref{sec:intro}), considering that fine-tuning still resulted in better adversarial test set performance in all cases, and since that would not require access to the original training data itself, it could be a feasible, easily implementable approach to improve the robustness of NER models without compromising on the original model performance. Re-training can be preferred when we have access to the original data and the model. Note that in both the cases, we are assuming no means to procure additional manually labeled training data, and the focus is on improving an NER model's robustness to adversarial input \textit{without} compromising on its performance on normal test data. 

\section{Conclusions and Discussion}
\label{sec:concl}
We explored simple, language agnostic approaches to generate adversarial test sets for NER and demonstrated their generalizability by testing on six datasets covering three languages - English, German and Hindi. 
While exact results differ depending on language/datasets, our key findings from these experiments can be summarized as follows:
\begin{enumerate}
    \item NER models for all three languages are sensitive to adversarial input.
    \item Adversarial fine-tuning and re-training could improve the performance of NER models both on original and adversarial test sets, without requiring additional manual labeled data.
\end{enumerate}  

The proposed approaches and tested languages/models are by no means comprehensive, and extending this work to include more NER models, adding new languages, and developing new adversarial data generation methods for NER is an obvious next step, as the current results provide enough evidence on the sensitivity of state of the art NER models to adversarial inputs. The methods we employed for adversarial fine-tuning/re-training too are just a starting point towards exploring the use of adversarial data in building more robust NER systems. We only explored one paraphraser for this task. The usefulness of the recent generative language models for creating such test data can be an interesting next step in this direction. 

\section{Limitations} The adversarial test sets based on masked language models can introduce new noise into the sentence context, as there is no way to automatically ensure grammatical correctness. However, there were many cases where such introduction of noise did not affect the predictions, in all three languages. Further, adversarial datasets are expected to introduce such noise, as is seen in other research on the topic for other tasks such as sentiment analysis, and the goal of such research is also to understand model robustness in the presence some noise. It is relevant to mention in this context that the NER datasets we considered already consist of other noise and ungrammatical examples such as score cards of sporting matches (conll03-en), social media content (wnut17) and fully lower-cased sentences with weakly supervised annotations (mconer21). Further, masking does not alter the entities themselves, and only changes the non-entity tokens. So, the NER models still see the same entities. While there are no established means of quantifying the quality of adversarial datasets to our knowledge, exploring human-in-the-loop approaches to select appropriate examples to include in the final adversarial test set can be one way to address the issue.  

\section{Ethics and Impact Statement}
The paper described the creation of several adversarial test sets for three languages. We used publicly available datasets for this purpose, and the research did not involve human participants. All the datasets we generated and the code to generate them are  shared as supplementary material\footnote{\url{https://dx.doi.org/10.6084/m9.figshare.22674079}}, for replication and to further this line of research. Our goal in this paper was to study the sensitivity of state of the art NER systems to adversarial data, and suggest ways to overcome it. As such, the generated datasets are expected to be used only for that purpose, and the limitations of current approaches are discussed in the previous section. Apart from this, since the paper focuses on more foundational question of evaluating NER systems in general, we do not foresee any other potential risks involved with this research. 

\paragraph{Broader Impact} Considering the number of practical usecases of NER across industries, and the growth of multilingual NLP, NER evaluation beyond English is more important than ever before. In this paper, we explored a previously unexplored space for Named Entity Recognition, i.e., evaluating NER systems beyond English for their sensitivity to adversarial input, which will hopefully lead into better evaluation strategies when developing NER systems across languages in future. 

\section*{Acknowledgements}
We thank Justin Lee and Gabriel Bernier-Colborne for their feedback on an earlier draft, and Edwin Thomas for feedback on the final draft of the paper. 
\bibliography{custom}
\bibliographystyle{acl_natbib}

\appendix

\section{Appendix}
\label{sec:appendix}

\subsection{Dataset Statistics: }
\label{sec:app-ds}
\begin{table}[!ht]
    \centering
    \begin{tabular}{|p{2.5cm}|l|l|l|}
    \hline
    Dataset & \# train & \# dev & \# test \\ \hline
    \multicolumn{4}{|c|}{\cite{Sang.Demeulder-03}}\\ \hline
    conll03-en & 14,987& 3,466 & 3,684 \\
    conll03-de & 12,705 & 3,068 & 3,160\\ \hline
    \hline wnut17\footnote{\url{https://huggingface.co/datasets/wnut_17}} & 3394 & 1009 & 1287\\ \hline \hline
    \multicolumn{4}{|c|}{\cite{malmasi-etal-2022-semeval}} \\ \hline
    mconer21-en  & 15,300&800 & 217,818\\
    mconer21-de & 15,300 & 800 &  217,824\\
    mconer21-hi & 15,300 & 800 &  141,565 \\ \hline
      \end{tabular} 
    \caption{Dataset statistics in terms of number of sentences per split}
    \label{tab:datasets}
\end{table}

We used the dev set from mconer21 to create adversarial test sets for all the three languages, considering the large size of its test set.  


\subsection{Detailed Evaluation}
\label{app:detailedeval}
\begin{table}[!ht]
    \centering
    \begin{tabular}{|l|l|l|l|l|}
    \hline
        ~ & \multicolumn{2}{c|}{orig. test set}& \multicolumn{2}{c|}{adv. test set}\\ \hline
        ~ & Strict & Exact & Strict & Exact \\ \hline
        Overall & 0.91 & 0.95 & 0.82 & 0.92 \\ \hline
        PER & 0.96 & 0.98 & 0.87 & 0.92 \\ \hline
        LOC & 0.92 & 0.96 & 0.85 & 0.95 \\ \hline
        ORG & 0.89 & 0.94 & 0.75 & 0.91 \\ \hline
        MISC & 0.83 & 0.89 & 0.79 & 0.88 \\ \hline
    \end{tabular}
    \caption{Strict versus Exact evaluation for English (conll-en)}
    \label{tab:se-eng}
\end{table}

\begin{table}[!ht]
    \centering
    \begin{tabular}{|l|l|l|l|l|}
    \hline
        ~ & \multicolumn{2}{c|}{orig. test set}& \multicolumn{2}{c|}{adv. test set}\\ \hline
        ~ & Strict & Exact & Strict & Exact \\ \hline
        Overall & 0.73 & 0.81 & 0.61 & 0.73 \\ \hline
        PER & 0.88 & 0.92 & 0.8 & 0.86 \\ \hline
        LOC & 0.78 & 0.83 & 0.67 & 0.76 \\ \hline
        PROD & 0.7 & 0.79 & 0.58 & 0.73 \\ \hline
        GRP & 0.68 & 0.81 & 0.55 & 0.71 \\ \hline
        CORP & 0.66 & 0.8 & 0.53 & 0.75 \\ \hline
        CW & 0.59 & 0.67 & 0.44 & 0.53 \\ \hline
    \end{tabular}
        \caption{Strict versus Exact evaluation for German (multiconer21-de)}
    \label{tab:se-deu}
\end{table}

\begin{table}[htb!]
    \centering
    \begin{tabular}{|l|l|l|l|l|}
     \hline
        ~ & \multicolumn{2}{c|}{orig. test set}& \multicolumn{2}{c|}{adv. test set}\\
        ~ & Strict & Exact & Strict & Exact \\ \hline
        Overall & 0.62 & 0.73 & 0.48 & 0.61 \\ \hline
        PER & 0.71 & 0.82 & 0.56 & 0.62 \\ \hline
        LOC & 0.77 & 0.82 & 0.57 & 0.72 \\ \hline
        PROD & 0.54 & 0.63 & 0.48 & 0.61 \\ \hline
        GRP & 0.67 & 0.81 & 0.52 & 0.67 \\ \hline
        CORP & 0.56 & 0.76 & 0.42 & 0.65 \\ \hline
        CW & 0.47 & 0.57 & 0.35 & 0.46 \\ \hline
    \end{tabular}
            \caption{Strict versus Exact evaluation for Hindi (mconer21-hi)}
    \label{tab:se-hi}
\end{table}

\begin{figure*}[htb!]
\centering
  \includegraphics[width=\linewidth]{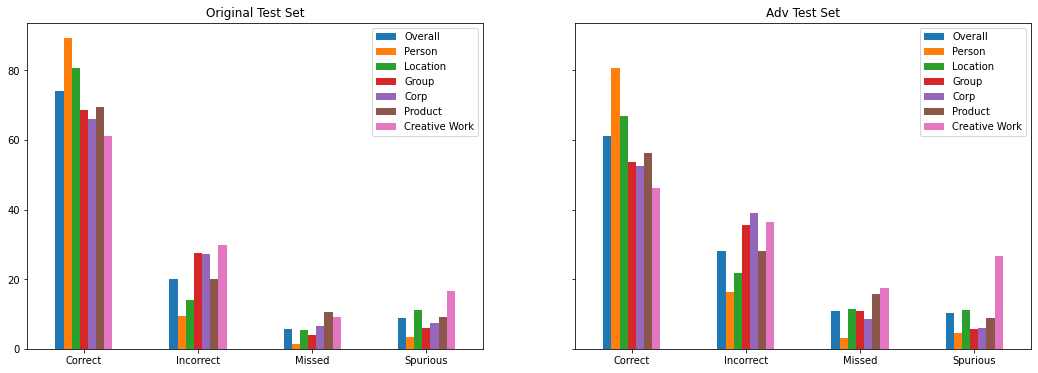}
  \caption{A plot visualising the performance of a German NER model (trained with mconer21 data) on original and M+R adversarial Test sets}
  \label{fig:plotde}
\end{figure*}

\begin{figure*}[htb!]
\centering
  \includegraphics[width=\textwidth]{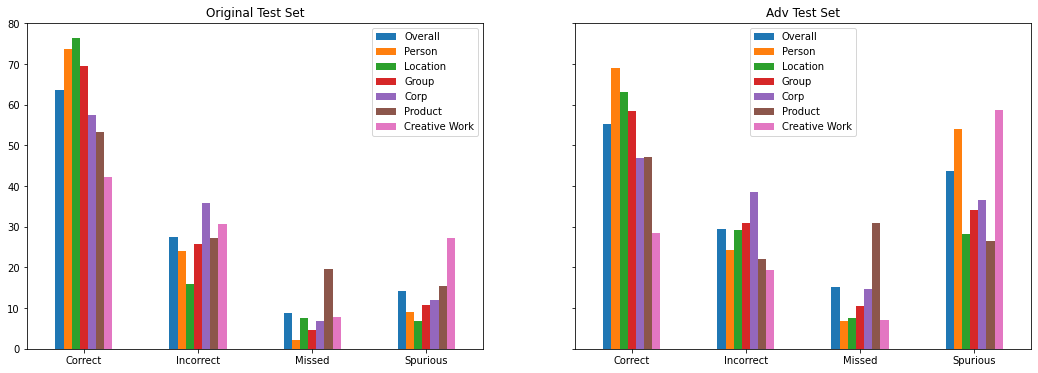}
  \caption{A plot visualising the performance of a Hindi NER model (trained with mconer21 data) on original and M+R adversarial Test sets}
  \label{fig:plothi}
\end{figure*}

\pagebreak\pagebreak
\begin{table*}[htb!]
\centering
\begin{tabular}{|p{15cm}|}
\hline \verb|Orig|: Der kleine \textbf{Elmir}$_{PER}$ verl\"aßt den Raum . \\ \hline
\hline \verb|RS|:\hspace{0.5cm} Der kleine \textbf{Treutel}$_{notidentified}$ verläßt den Raum .\\ 
\hline \verb|Faker|: Der kleine Melek$_{notidentified}$ verläßt den Raum . \\ \hline\hline
\verb|Orig|: Die Verwertungsgesellschaft Gebrauchte Kunststoffverpackungen in [\textbf{Bad Homburg}]$_{LOC}$ sei " offenbar mit ihrer Aufgabe \"{u}berfordert " . \\ \hline
\verb|RS|: Die Verwertungsgesellschaft Gebrauchte Kunststoffverpackungen in [\textbf{Stadt}]$_{LOC}$  [\textbf{Hanau}]$_{notidentified}$ sei " offenbar mit ihrer Aufgabe \"{u}berfordert " . \\ \hline\hline
\verb|Orig|: Im [Korea-Krieg]$_{MISC}$ \textbf{hatte} China$_{LOC}$ das kommunistische Nordkorea$_{LOC}$ \textbf{unterst\"{u}tzt} \textbf{.} \\ \hline
\verb|Mask|: Im Korea-Krieg$_{LOC}$ \textbf{hatten} China$_{LOC}$ das kommunistische Nordkorea$_{LOC}$ \textbf{gewonnen} \textbf{$|$} \\ \hline\hline
\verb|Orig|: Ihm stehen 20 000 Mark zur \textbf{Verfügung} , um \textbf{die Ausstellung} im November 1993 im Stadtmuseum$_{LOC}$ zu realisieren . \\ \hline
\verb|Mask|: Ihm stehen 20 000 Mark zur \textbf{wahl} , um \textbf{diese skulptur} im November 1993 im Stadtmuseum$_{notidentified}$ zu realisieren . \\ \hline \hline
\verb|Orig|: \textbf{Oder die} [Gauck-Beh\"orde]$_{MISC}$ ? \\ \hline
\verb|Mask|: \textbf{in der} Gauck-Beh\"orde]$_{LOC}$ ? \\ \hline
\end{tabular}
\caption{Examples of cases where a German NER model fails on adversarial input, but makes correct predictions on the original text}
\label{tab:erroranalde}
\end{table*} 

\begin{figure*}[htb!]
    \includegraphics[width=\textwidth]{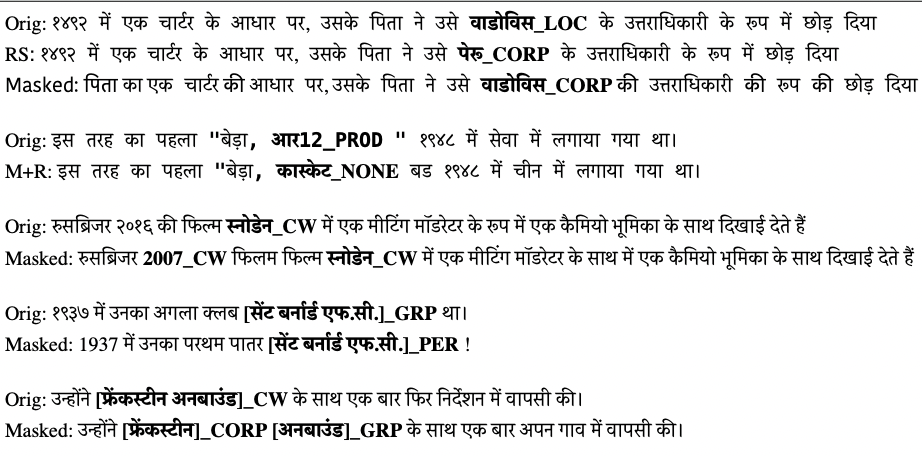}
    \caption{Examples comparing how predictions change for a Hindi NER model on original versus adversarial test sets}
    \label{fig:erroranalhi}
\end{figure*}

\end{document}